\documentclass{article}

\usepackage[nonatbib,final]{neurips_2023}

\usepackage[utf8]{inputenc} 
\usepackage[T1]{fontenc}    
\usepackage[colorlinks=true, linkcolor=black, citecolor=black, urlcolor=black]{hyperref}       
\usepackage{url}            
\usepackage{booktabs}       
\usepackage{amsfonts}       
\usepackage{nicefrac}       
\usepackage{microtype}      
\usepackage{xcolor}         

\usepackage{mathtools}
\usepackage{amsthm,bm}
\usepackage{algorithm} 
\usepackage{algorithmic}
\usepackage{times}
\usepackage{helvet}
\usepackage{courier}
\usepackage[caption=false]{subfig}
\usepackage{soul}
\usepackage{array}
\usepackage[backend=biber]{biblatex}
\usepackage{scalerel}

\usepackage[toc,page]{appendix}

\DeclareMathOperator*{\concat}{\scalerel*{\Vert}{\sum}}
\def\BibTeX{{\rm B\kern-.05em{\sc i\kern-.025em b}\kern-.08em
    T\kern-.1667em\lower.7ex\hbox{E}\kern-.125emX}}
\usepackage{tikz}

\title{One Node Per User: Node-Level Federated Learning for Graph Neural Networks}

\author{%
    Zhidong Gao\textsuperscript{\rm 1}, 
    Yuanxiong Guo\textsuperscript{\rm 2}, 
    Yanmin Gong\textsuperscript{\rm 1}\textsuperscript{\dag}\\
   \textsuperscript{\rm 1}Department of Electrical and Computer Engineering\\
   \textsuperscript{\rm 2}Department of Information Systems and Cyber Security\\
  The University of Texas at San Antonio\\
  San Antonio, USA \\
  \{zhidong.gao@my., yuanxiong.guo@, yanmin.gong@\}utsa.edu \\
}

\begin{document}

\maketitle
\renewcommand{\thefootnote}{\dag}
\footnotetext{Corresponding Authors}

\begin{abstract}
Graph Neural Networks (GNNs) training often necessitates gathering raw user data on a central server, which raises significant privacy concerns. Federated learning emerges as a solution, enabling collaborative model training without users directly sharing their raw data. However, integrating federated learning with GNNs presents unique challenges, especially when a client represents a graph node and holds merely a single feature vector. In this paper, we propose a novel framework for node-level federated graph learning. Specifically, we decouple the message-passing and feature vector transformation processes of the first GNN layer, allowing them to be executed separately on the user devices and the cloud server. Moreover, we introduce a graph Laplacian term based on the feature vector's latent representation to regulate the user-side model updates. The experiment results on multiple datasets show that our approach achieves better performance compared with baselines.
\end{abstract}

\section{Introduction}
Graph Neural Networks (GNNs) have attracted significant attention both within academic circles and across diverse industries. Their remarkable achievements span a multitude of domains, including fraud detection in social networks~\cite{weber2019anti}, cancer classification in biology science~\cite{rhee2017hybrid}, and materials design in molecular chemistry~\cite{duvenaud2015convolutional}. In real-world applications, graph data tied to individuals or human behaviors often contains sensitive details. For example, the user's comments, friend list, and profiles on a social platform, as well as their purchase records, browsing history, and transactions on an economic network, are typically deemed private. With increasing emphasis on user privacy, legal restrictions like the General Data Protection Regulation (GDPR) in Europe (EU) and the Health Insurance Portability and Accountability Act (HIPAA) in the US have rendered data-sharing practices infeasible. However, the conventional graph machine learning paradigm requires uploading raw user data to a central server. This is infeasible due to privacy restrictions, hindering the deployment of many real-world graph-based applications.

Federated learning (FL) \cite{mcmahan2017communication} offers a collaborative learning method, allowing multiple clients to train a model without revealing their raw training data. The workflow of FL follows an iterative training procedure, including multiple communication rounds between the clients and a central server. Specifically, the central server maintains a global model and orchestrates the training process. In each round, the selected clients fetch the global model and perform several epochs of updating using their local training data. The updated local model is later uploaded to the server to produce the latest global model. The training process terminated until the model meets the pre-defined criteria. 

Some efforts have recently been devoted to training GNN models over graph data with the preservation of user privacy. One straightforward approach is extending FL to graph machine learning. Depending on the available data possessed by the clients, these research studies  \cite{wang2020graphfl,he2021fedgraphnn,yao2022fedgcn,hu2022fedgcn,wu2021fedgnn,chen2020vertically} assume either graph-level FL, where multiple graphs are distributed among clients, or subgraph-level FL, where a big graph is partitioned into several subgraphs and each client has access to the subgraph. With at least one (sub)-graph possessed by the client, local training could proceed as standard FL algorithm plus certain adaptations (e.g., missing link retrieve) tailored for graph data. Note these methods fail under node-level data availability, where each client only possesses data of one node locally (one feature vector). Another line of research work \cite{sajadmanesh2021locally} aims to mitigate the privacy risk by introducing Differential-Private (DP) noise to user data, and the model training process follows the conventional machine learning paradigm. As the added DP noise inevitably degrades the quality of training data, this approach suffers from an inherent privacy-performance trade-off. Consequently, the potential of the GNNs to achieve better performance is restricted. In summary, fewer prior works (see Appendix~\ref{sec:related} for detailed related work) have explored training GNNs under node-level user data privacy protection. 

\begin{figure*}
  \centering
    \includegraphics[width=0.9\textwidth]{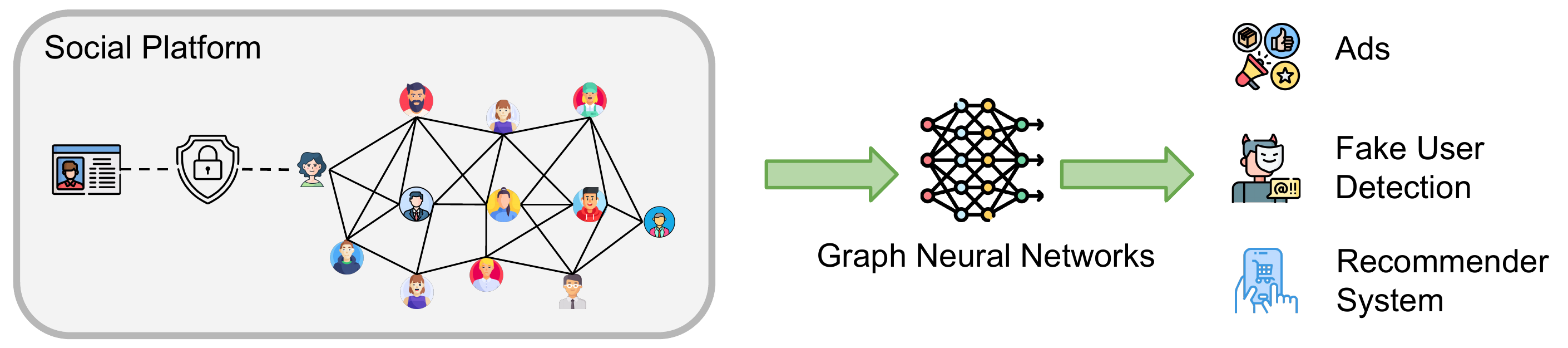}
  \caption{Node-level collaborative training framework for GNNs: a motivating example.
  }\label{fig:motivation}
\end{figure*}

\textbf{Motivating Example.} 
As depicted in Fig.~\ref{fig:motivation}, our motivating example showcases how industries often use graphs wherein each node represents an individual user. Take mobile social networks as an instance: here, a node represents a mobile user, and an edge signifies the social ties between them. These platforms are interested in using additional user data in user's mobile devices, such as users' locations, activity records, and app usage histories, to perform tasks such as Sybil detection \cite{shu2017fake}, online advertisements \cite{li2019fi}, and recommendations of social media content \cite{guo2020deep}. Similarly, telecom giants such as AT$\&$T and Spectrum possess vast amounts of phone call logs between users. These logs can be depicted as a graph where each node symbolizes a user, and an edge indicates a call record shared between two users. To enhance user experience, these companies might desire access to richer user datasets, such as users' app usage statistics, geolocation data, and device sensor information. Such details are instrumental in understanding user preferences and needs.

\textbf{Challenges.} Implementing node-level FL within GNNs presents several unique challenges. Foremost, each user's training data is remarkably limited, consisting merely of one feature vector. This is different from existing FL settings, where each client has enough data (a graph or sub-graph) to train a local model. Moreover, the graph-based node classification task belongs to the transductive learning (semi-supervised learning) regime, wherein the unlabeled nodes also contribute to the model training. However, conventional FL algorithms such as FedAvg \cite{mcmahan2017communication} inherently assume user data to be labeled (or partially labeled). It is untrivial to integrate the knowledge from unlabeled nodes (clients) into the GNNs under the node-level FL setting.

In this paper, we introduce a node-level FL framework for GNNs, termed~\textbf{nFedGNN}. This framework facilitates collaborative training across multiple clients, with each client only possessing data of a single node\footnote{We will use client, user, and node interchangeably in this paper when it does not introduce ambiguity.} in the graph, specifically, one feature vector. Importantly, this feature vector remains localized throughout the training, ensuring no external disclosure. Moreover, we assume that the server has labeled some nodes, and the learning objective is to assign the label for the unlabeled nodes. We also assume the central server has access to the graph topology. Note the research community usually believes the graph topology is private in graph-level and subgraph-level FL. However, under node-level data availability of the client, it is unnecessary to consider the privacy of the graph topology. As illustrated in the motivating example, this assumption aligns with many real-world GNN applications.

\textbf{Key ideas of our solution.} 
To facilitate GNN training across distributed users, our method involves splitting the GNN model between the clients and the server, similar to the technique used in split learning or vertical federated learning \cite{singh2019detailed, vepakomma2018split, chen2020vafl}. Specifically, the user-side model processes the local private feature vector, yielding a latent representation vector. Subsequently, the server gathers these latent vectors and integrates them with graph topology through the server-side model. To update the client-side model, the server computes the gradient of the latent representation vectors and transmits it back to the clients. The entire training process needs multiple communication rounds until the model meets the predefined criteria. We conducted a preliminary experiment, and the result shows this approach is prone to overfitting, leading to subpar generalization performance.

We analyze the potential causes and conclude that the user-side model of labeled nodes easily fits their local data. To address this, we introduce an explicit graph regularization loss $L_{\text{reg}}$ based on the received latent vectors. Our solution is motivated by the fundamental assumption of graph-based semi-supervised learning: connected nodes in a graph are likely to share the same label \cite{zhu2003semi, weston2008deep}. This implies that connected nodes are likely to share a similar node embedding. In the preliminary experiment, this assumption is broken as each user-side model has enough flexibility in its embedding. The regularization loss imposes constraints on the distribution of latent representations, effectively limiting the degree of freedom in the latent space. Consequently, the user-side model is not only required to fit its own local data but also has a strong motivation to learn from its neighbors. As a result, the learned model achieves better performance. 
We conducted the experiment on several datasets using split GCN and GAT models. The experiment result shows that nFedGNN significantly surpasses the baselines. The contributions of this paper are summarized as follows:

\begin{itemize}
\item We propose a node-level FL framework for training GNNs, where each participating user only has access to the data of a single node. The framework inherits the privacy advantage of FL and enables efficient utilization of distributed graph data. Our approach is one of the very few early works in this field, and we are the first to provide a general FL-based approach in this setting. 

\item We instantiate two popular GNN models, GCN and GAT, within our proposed framework. For both models, the first layer is divided into the server and the users. The remaining layers reside on the server side, functioning as the conventional GCN and GAT models do.  Our framework allows us to train the cutting-edge GNN model and achieve better performance without collecting the user's data.

\item We perform extensive experiments to evaluate the proposed framework on multiple datasets. The results show that the proposed method outperforms the baselines on all datasets for both GCN and GAT models.
\end{itemize}

\begin{figure}
  \centering
    \includegraphics[width=1.0\linewidth]{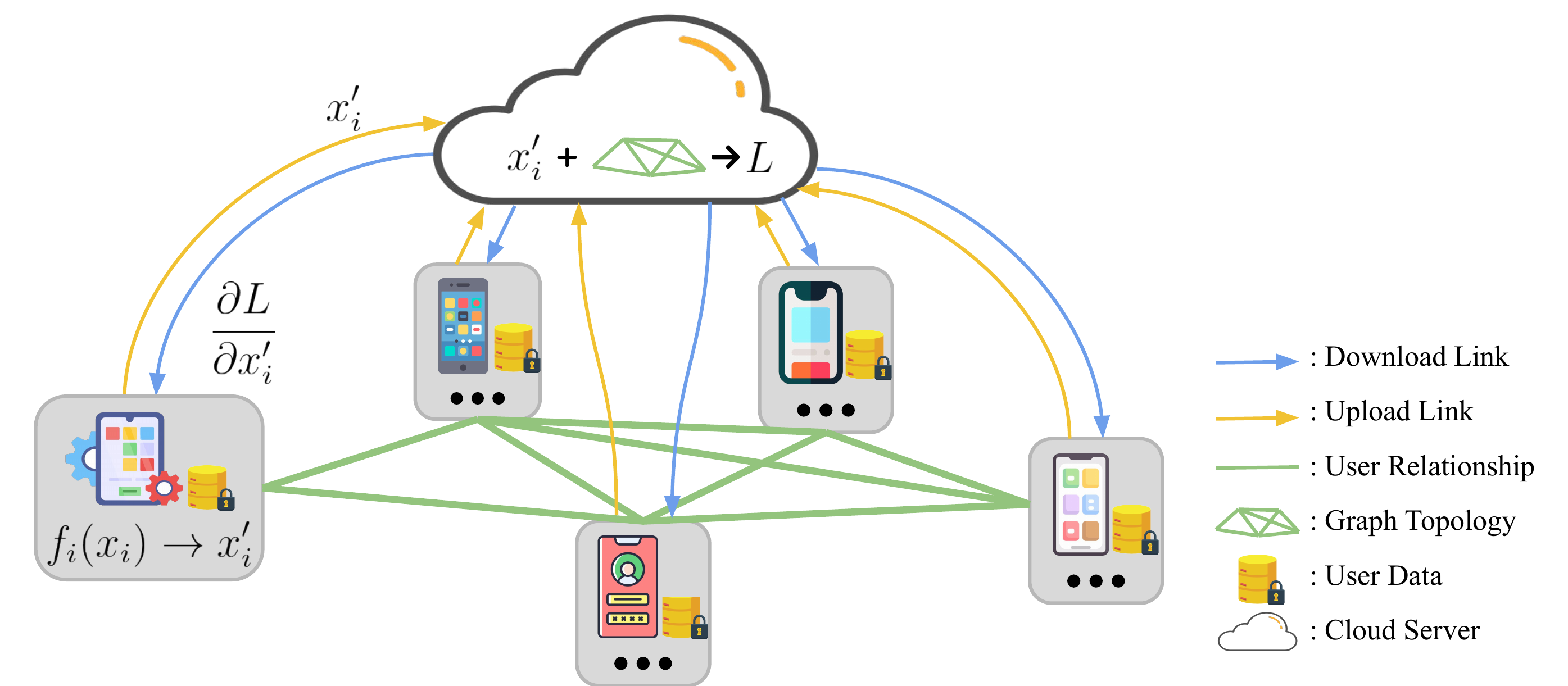}
  \caption{An illustration of the proposed node-level FL framework nFedGNN. Here, each node of the graph is a client of the FL system. The server knows the topology of the graph and the label of a small subset of nodes. The server has no access to the user's raw data while it wants to train a GNN model to assign the label for unlabeled nodes. 
  }\label{fig:sys}
\end{figure} 

The remainder of this paper is organized as follows. In 
Section \ref{sec:problem}, we present details of the GNN and FL system. In Section \ref{sec:solution}, we describe the proposed algorithm nFedGNN. Then, in Section \ref{sec:exp}, we demonstrate experiment results and analysis. Finally, we conclude the paper and discuss the future direction in Section \ref{sec:con}.

\section{Preliminary}\label{sec:problem}

\subsection{Graph Neural Networks}
Let $\mathcal{G} = (\mathcal{V}, \mathcal{E}, X)$ be an attributed graph, where $\mathcal{V}$ is set of nodes, and $\mathcal{E}$ is set of edges. The graph topology can be represented as an adjacency matrix $A \in \mathbb{R}^{n \times n}$, $A_{i, j} = 1$ if there is an edge $(i, j) \in \mathcal{E}$ between node $i$ and node $j$, and $A_{i, j} = 0$ otherwise. Here $n = |\mathcal{V}|$ denotes the number of nodes. $X \in \mathbb{R}^{n\times d}$ is the feature matrix, where each row corresponds to a node's feature vector $x_{i}\in \mathbb{R}^d$. Denote $Y \in \mathbb{R}^{n\times c}$ as a matrix of the one-hot labels, where each row corresponds to a one-hot vector of a labeled node and $c$ is the number of classes. Note only a subset of the nodes $\mathcal{V}_0 \subset \mathcal{V}$ are labeled.

Given a graph dataset $\mathcal{G}$, the $K$-layer GNN consists of $K$ consecutive layers, where each layer $k$ receives as input the node embeddings $\{\bm{h}_i^{k-1}, \forall i \in \mathcal{V}\}$ from layer $k-1$ and outputs a new node embedding $\bm{h}_i^k$ for each node $i$ by aggregating the current embeddings of its adjacent neighbors followed by a learnable transformation as follows:
\begin{equation}\label{equ:messpass}
    \begin{split}
        \bm{h}^{k}_{\mathcal{N}_i} & := \textbf{Agg}_k\left(\{\bm{h}_j^{k-1}, \forall j \in \mathcal{N}_i\}\right),\\
\bm{h}^k_{i} & :=  \textbf{Update}_k\left(\bm{h}^{k}_{\mathcal{N}_i}\right),
    \end{split}
\end{equation}
where $\mathcal{N}_i$ is the set of neighbors of node $i$, including itself, in the graph $\mathcal{G}$. Here $\textbf{Agg}_k(\cdot)$ is the aggregator function (e.g., mean, sum, and max) for $k$-th layer, and $\textbf{Update}_k (\cdot)$ is the $k$-th layer trainable non-linear function (e.g., neural network) for $k$-th layer. The initial embedding of each node $i$ is its feature vector, i.e., $\bm{h}^0_i = \bm{x}_i$, and the node embeddings from the last layer's output $\{\bm{h}_i^K, \forall i \in \mathcal{V}\}$ will be used for downstream tasks such as predicting the labels of nodes from the unlabeled set $\mathcal{V}\setminus \mathcal{V}_0$.  
\subsection{Problem Definition}

In this paper, we explore learning a GNN model under the node-level federated graph learning setting, where each node in the graph represents a user, and the node feature $x_i$ is private to user $i$. Furthermore, we assume a central server has access to the graph topology $A$ as well as the label of the node in $\mathcal{V}_0$, but can not observe the feature matrix $X$. The central server aims to cooperate with the users to train a GNN model over the graph $\mathcal{G}$ without requiring the private data $x_i$ to leave the users. We focus on the node classification task, where the server wants to assign labels to the remaining unlabeled nodes (users) $\mathcal{V}\setminus \mathcal{V}_0$ in the graph. 
\subsection{FL Objective}
Let $f_s(\theta^s, A, \cdot)$ be the server-side model, parameterized by $\theta^s$, and $f_u(\theta_i^u, \cdot)$ be the user-side model, parameterized by $\theta_i^u$. The user-side model's output is $\bar{x}_i=f_u(\theta_i^u,x_i)$. Then, the GNN model's final output is:
\begin{equation}\label{equ:infe}
\begin{split}
     Z  & = f_{s}(\theta^{s}, A, [\bar{x}_1, \cdots, \bar{x}_n]^{T}) \\
     & = f_{s}(\theta^{s}, A, [f_{u}(\theta_{1}^{u},x_{1}), \cdots, f_{u}(\theta_{n}^{u},x_{n})]^{T}). 
\end{split}  
\end{equation}
Define $\theta = \{\theta^{s}\} \bigcup \{\theta_i^{u}\}_{i=1}^{n}$ be the set of all trainable parameters on both user-side and server-side. Then, the goal of GNN training under node-level FL is to minimize the following classification loss:
\begin{equation}\label{equ:opt_ce}
    \min_{\theta} \mathcal{L}_{\text{CE}}(\theta) := \sum_{l \in \mathcal{V}_0} \text{CE}(Z_{l}(X,\theta), Y_{l}). 
\end{equation}
Here, $\text{CE}$ denotes the cross-entropy loss, and $\mathcal{V}_0$ is the set of labeled nodes. 

\section{The Proposed Method}\label{sec:solution}
\subsection{Algorithm to address (\ref{equ:opt_ce}).}
\begin{algorithm}[tb]
    \caption{nFedGNN}
    \label{alg:nFedGNN}
\begin{algorithmic}[1]
    \STATE {\bfseries Input:} training data $X =[x_{1}, x_{2}, \cdots, x_{n}]^{T}$
    \STATE {\bfseries Output:} learned model $\theta = \{\theta^{s}\} \bigcup \{\theta_i^{u}\}_{i=1}^{n}$
    \FOR{round $t = 1,2,\ldots, T$}
        \FOR{user $i = 1,2,\ldots,n$, \textbf{in parallel}}
            \STATE Compute the output of user-side model: $x_{i}' = f_{u}(\theta_{i}^{u},x_{i})$. \label{ln:1}
            \STATE Upload $x_{i}'$ to the server. \label{ln:2}
        \ENDFOR
        \STATE Server updates $\theta^{s}$ through gradient decent. \label{ln:5}
        \STATE Server computes $\frac{\partial \mathcal{L}_{CE}}{\partial \bar{x}_i}$ and send it back to user. \label{ln:6}
        \FOR {each user $i = 1,2,\ldots,n$, \textbf{in parallel}}
            \STATE Compute the gradient for $\theta_{i}^{u}$ using $\frac{\partial \mathcal{L}}{\partial x_{i}'}$. \label{ln:7}
            \STATE Update $\theta_{i}^{u}$ through gradient decent.\label{ln:8}
        \ENDFOR
    \ENDFOR
\end{algorithmic}
\end{algorithm}
We summarize the details for addressing (\ref{equ:opt_ce}) in \textbf{Algorithm \ref{alg:nFedGNN}}. Specifically, at the beginning of the $t$-th communication round, every user (node) in the graph computes the latent output of the user-side model and uploads $x_{i}'$ to the server (lines~\ref{ln:1}-\ref{ln:2}). Subsequently, the server computes the classification loss based on equations (\ref{equ:infe}) and (\ref{equ:opt_ce}). Following this, the server-side model parameter $\theta^{s}$ can be updated using standard back-propagation and gradient descent (lines~\ref{ln:5}). The server then computes the gradient for the user-side model's output $\frac{\partial \mathcal{L}_{CE}}{\partial \bar{x}_i}$ and sends it to the corresponding user (line~\ref{ln:6}). The user receives $\frac{\partial \mathcal{L}_{CE}}{\partial \bar{x}_i}$ and computes the gradient of the user-side model through back-propagation (line~\ref{ln:7}). Finally, the user-side model parameter $\theta_{i}^{u}$ is updated via gradient descent (line~\ref{ln:8}). 

\subsection{Model Splitting Strategy}
Implementing a layer-wise split of the GNN model is unfeasible in a node-level FL setting as the resultant user-side model still requires the entire graph as input. To address this problem, we start from the message-passing mechanism of the GNN layer and suggest splitting the $\textbf{Agg}(\cdot)$ and $\textbf{Update}(\cdot)$ functions of the first GNN layer. A detailed discussion of the general model splitting process can be found in Appendix~\ref{app:mps}. In this section, we elucidate this concept through a simplified example of a two-layer GCN model.

Let $\hat{A} = \tilde{D}^{-\frac{1}{2}} \tilde{A} \tilde{D}^{-\frac{1}{2}}$ be the normalized adjacency matrix, where $\tilde{A} = A+I$ is the adjacency matrix with self-connections, $\tilde{D}_{ii}=\sum_{j}\tilde{A}_{ij}$ is the degree matrix, and $I\in \mathbb{R}^{n \times n}$ is the identity matrix. Then, the forward pass of the standard 2-layer GCN model \cite{kipf2016semi} can be expressed as follows
\begin{equation}
Z = \text{softmax }(\tilde{A}\text{ ReLU}(\tilde{A}XW^{(0)})W^{(1)}),
\end{equation}
where $W^{(0)}$ and $W^{(1)}$ are the learnable weight matrix of the first and second layers, respectively. 

Following the proposed model splitting strategy, we split the first GCN layer into two parts: every user retains a local $W^{(1)}$, while $\tilde{A}$ is located on the server side. The training process of the split GCN model unfolds as follows. In each round, the user-side model first maps its feature vector $x_i$ into latent representation $\bar{x}_i=xW_i^{(0)}$, where $W_i^{(0)}$ is the weight matrix of the user $i$. Subsequently, $x_i$ is uploaded to the server. The server-side model computes the loss based on the topology and the received latent vectors. Let $\bar{X} = [\bar{x}_1,\dots,\bar{x}_n]^T$. The forward pass of the server-side model is expressed as:
\begin{equation}\label{equ:spit-gcn}
Z = \text{softmax }(\tilde{A}\text{ReLU}(\tilde{A}\bar{X})W^{(1)})
\end{equation}

\subsection{Preliminary Evaluation}
\begin{figure}[t]
\centering
    \subfloat[]{{\includegraphics[width=0.4\textwidth]{ {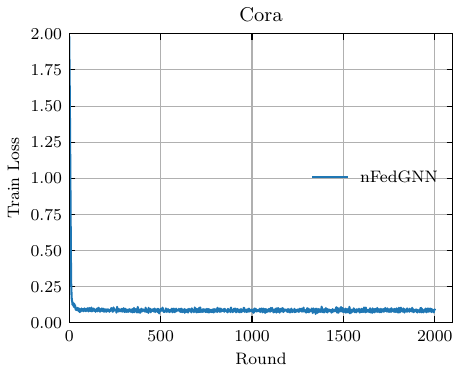}} }}
    \subfloat[]{{\includegraphics[width=0.4\textwidth]{ {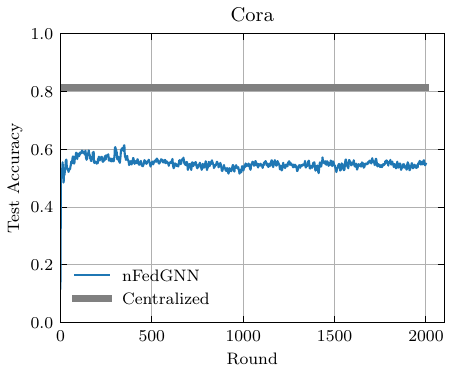}} }} \\
    \caption{Training curves of a split two-layer GCN on the Cora dataset. Here, we run the experiment with three random seeds and report the average result. The grey line shows the accuracy of the centralized GCN: 81.5 \% \cite{kipf2016semi}.}
    \label{fig:2000_fedgcn}%
\end{figure}

We train the split GCN model on the Cora dataset \cite{sen2008collective} to assess the performance of the~\textbf{nFedGNN}. The training curves are shown in Fig.(\ref{fig:2000_fedgcn}). From the figure, we can observe that the training loss quickly decreases after only a few rounds of updating and then does not change much, and improvement in testing accuracy is only seen in the first few rounds. We also plot the testing accuracy of the GCN in the centralized training setting~\cite{kipf2016semi} as reference. Although the proposed method works, the final testing accuracy still leaves room for improvement.

We examine the internal state of the GNN model after splitting. In nFedGNN, we instantiate a local model for every node in the graph. For the user who possesses the labeled node, their local model only undergoes minor adjustments (as the local data is extremely scarce) to fit the local feature vector during training. Consequently, the local model of a labeled user rapidly adapts to their local data, diminishing the training loss to a near-zero value. However, the user-side model from the unlabeled nodes still remains un-updated.

\subsection{nFedGNN with regularization}\label{sec:nFedGNN}
To improve the performance of nFedGNN, we revisited the foundational assumption of graph-based semi-supervised learning: the connected nodes in the graph are likely to share the same label \cite{zhu2003semi, weston2008deep}. If two nodes are connected in the graph, a properly trained GNN model is likely to assign the same label to them. This suggests that the latent representations of two connected nodes should become increasingly similar as the model's layers become deeper, even if they have distinct feature vectors. 

Drawing from this observation, in the context of the split GNN model, if two users are interconnected, their respective latent representations should ideally manifest analogous patterns. However, as discussed previously, the user-side model is only responsible for one feature vector locally. Consequently, the node's latent representation shares less similarity with neighbor nodes. If we restrict the distribution of latent representations and let the connected users have a similar latent representation, the performance of the learned model should be improved. Motivated by this, we introduce the graph Laplacian regularization based on the received latent representations:
\begin{equation}\label{equ:reg_loss}
    \mathcal{L}_{\text{reg}} = \frac{1}{ \sum_{i=1}^{n}|\mathcal{N}_{i}|}\sum_{i=1}^{n}\sum_{j=1}^{\mathcal{N}_{i}}\|f_{u}(\theta_{i}^{u},x_{i})-f_{u}(\theta_{j}^{u},x_{j})\|^2. 
\end{equation}
The overall training loss can be formulated as
\begin{equation}\label{equ:loss}
\mathcal{L} = \mathcal{L}_{CE} + \lambda \mathcal{L}_{\text{reg}}. 
\end{equation}
Here, $\lambda \geq 0$ is the hyperparameter that balances the weight of classification loss and regularization. 

\subsection{Discussion}
\textbf{Graph Regularization:} Graph regularization-based approach has a long history and is widely used for various applications. It’s noteworthy that the state-of-the-art GNN models \cite{kipf2016semi} relax the assumption behind the regularization-based approaches, where the connected nodes in the graph are likely to share the same label. Instead, they directly encode the topology with the feature matrix by a neural network, as the graph topology does not necessarily encode similarity but contains the information that does not appear in the feature matrix $X$. Nevertheless, this assumption still holds for a large portion of nodes within the graph, a fact evidenced by the success of regularization-based strategies. In nFedGNN, we use the graph Laplacian regularization to restrict the distribution of latent representation from the user-side model. The remaining layers of the GNN model (located on the server side) still encode the graph topology with the (latent)-feature matrix as the state-of-the-art GNN model. Consequently, the model still utilizes the knowledge from graph topology as state-of-the-art GNNs.

\textbf{Communication Cost:} In cross-device federated learning environments, the communication bottleneck is one of the major concerns in prior graph-level and subgraph-level FL researches~\cite{yao2022fedgcn, kairouz2021advances}. However, within nFedGNN, the data exchanged between the user and server each round consists solely of a latent representation vector and a corresponding gradient vector, as opposed to the entire model parameter. Typically, the length of the latent representation vector is in the range of tens to hundreds depending on the tasks and models, substantially smaller compared to the total count of neural network parameters. Thus, within the framework of nFedGNN, communication cost does not pose as a constraining factor.
\section{Experiments}\label{sec:exp}
\begin{figure}[t]
\centering
    \subfloat[]{{\includegraphics[width=0.33\textwidth]{ {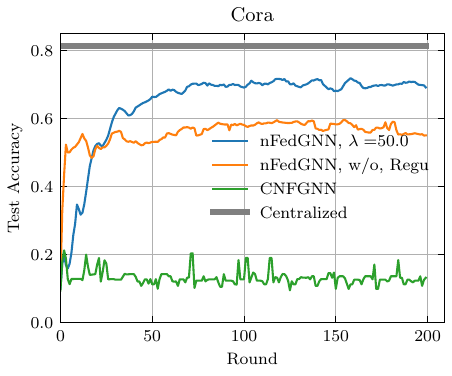}} }}
    \subfloat[]{{\includegraphics[width=0.33\textwidth]{ {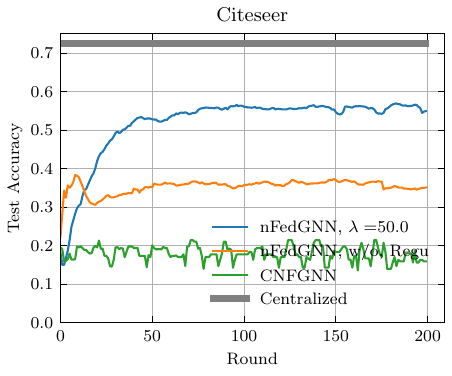}} }}
    \subfloat[]{{\includegraphics[width=0.33\textwidth]{ {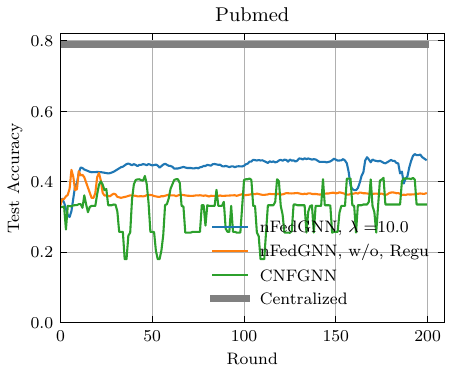}} }} \\
    \subfloat[]{{\includegraphics[width=0.33\textwidth]{ {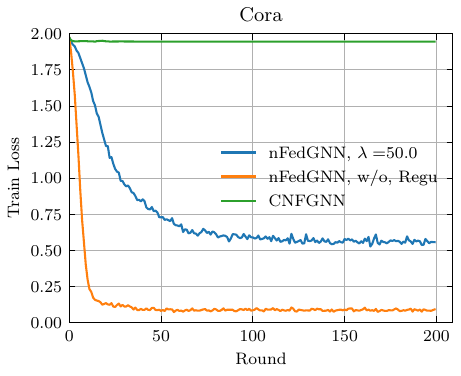}} }}
    \subfloat[]{{\includegraphics[width=0.33\textwidth]{ {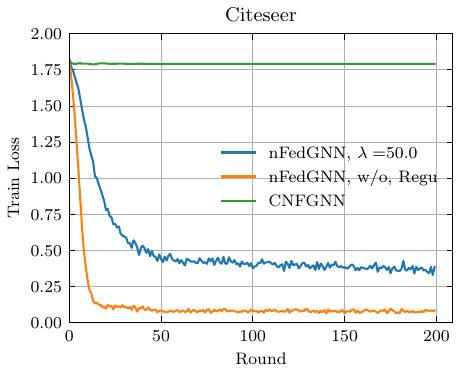}} }}
    \subfloat[]{{\includegraphics[width=0.33\textwidth]{ {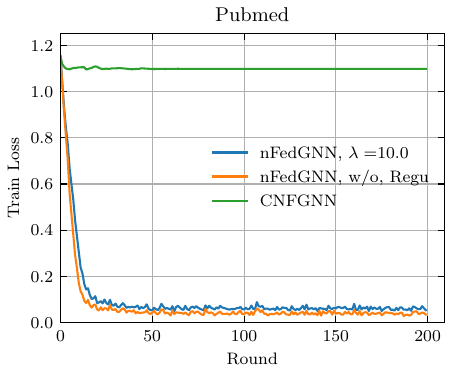}} }}
    \caption{nFedGNN vs baseline in terms of testing accuracy and training loss of a two-layer GCN on the Cora, Citeseer, and Pubmed datasets.}%
    \label{fig:compare_baseline_gcn}%
\end{figure}

\begin{figure}[t]
\centering
    \subfloat[]{{\includegraphics[width=0.33\textwidth]{ {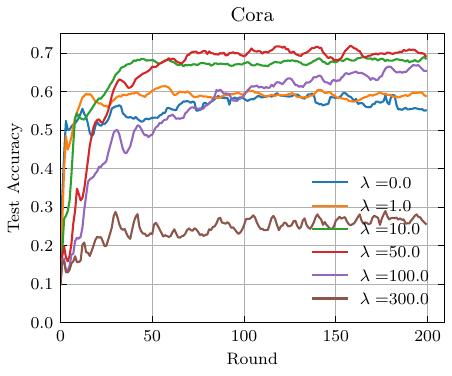}} }}
    \subfloat[]{{\includegraphics[width=0.33\textwidth]{ {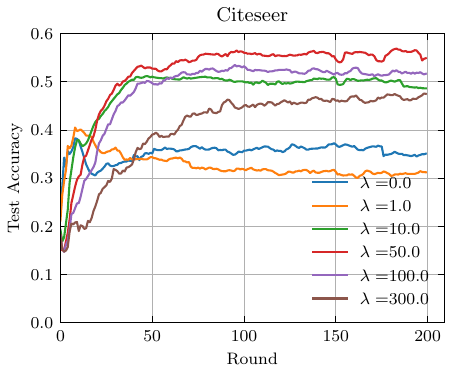}} }}
    \subfloat[]{{\includegraphics[width=0.33\textwidth]{ {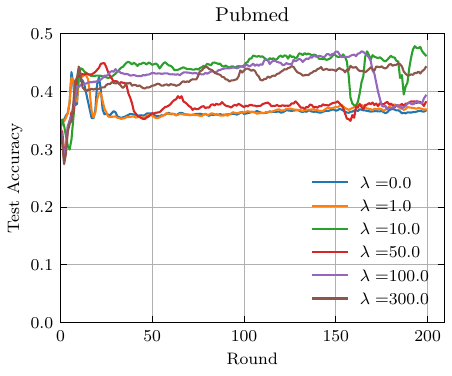}} }} \\
    \subfloat[]{{\includegraphics[width=0.33\textwidth]{ {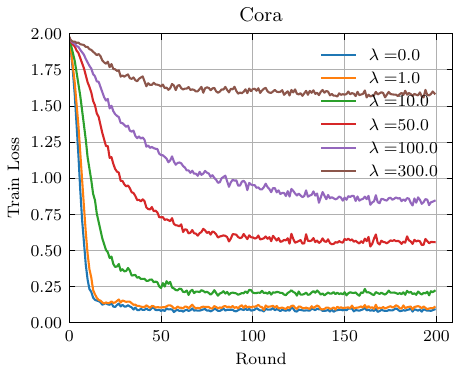}} }}
    \subfloat[]{{\includegraphics[width=0.33\textwidth]{ {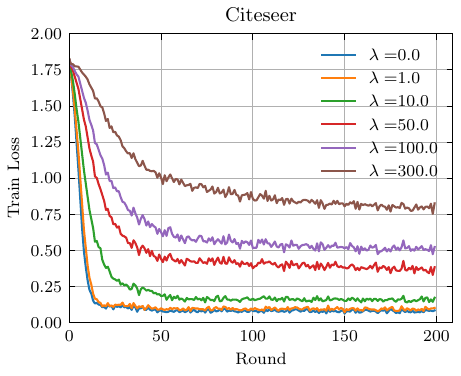}} }}
    \subfloat[]{{\includegraphics[width=0.33\textwidth]{ {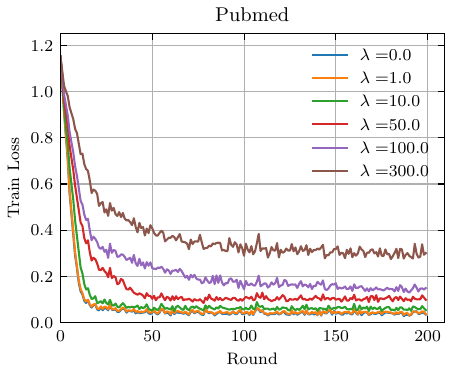}} }}
    \caption{Impact of $\lambda$ of nFedGNN on the Cora, Citeseer, and Pubmed datasets. }%
    \label{fig:lambda_gcn}%
\end{figure}

\subsection{Experimental Setup}
We evaluate nFedGNN on six datasets: Cora, Citesser, Pubmed \cite{sen2008collective}, Chameleon, Squirrel~\cite{rozemberczki2021multi}, and Wiki-CS~\cite{mernyei2020wiki}. The details and statistics of these datasets can be found in Appendix~\ref{sec:data_stat}. Currently, we assume all users participate in the learning process at every communication round. The algorithm was implemented using DGL \cite{wang2019dgl} with the Pytorch backend, and the training/testing data partition is the same as prior works~\cite{kipf2016semi,velivckovic2017graph,pei2019geom,mernyei2020wiki}. All experiments were conducted on a GPU server with 4 NVIDIA RTX A6000 (48GB GPU memory). 

We employ two well-established GNN models -- GCN\cite{kipf2016semi} and GAT\cite{velivckovic2017graph} -- for the node classification task across all datasets. The model splitting strategy for GCN follows the strategy discussed in Section~\ref{sec:problem}. For GAT, a two-layer model is utilized, and the details of the model splitting strategy are discussed in Appendix~\ref{sec:gat_split}. To ensure a fair comparison, we let the model structure be the same as the model in centralized training counterparts before splitting. Specifically, the GCN model parameters are as follows: 16 hidden neurons, 0.5 dropout rate, and a learning rate of 0.1. For GAT, the dropout rate is 0.6, the number of attention heads is 8, the number of hidden neurons is 16, and the learning rate is 0.01. For both models, we employ the Adam optimizer with weight decay as $5e^{-4}$ to update the model parameter both on the server side and the user side. The total FL round number is set to be 200 for all experiments. We run each experiment with three random seeds and report the averaged training loss and testing accuracy. 

\textbf{Baseline:} Given limited exploration in prior studies regarding node-level federated graph learning, establishing a baseline for comparison is challenging. A study close to ours is \textbf{CNFGNN}~\cite{meng2021cross}, which advocates for the transmission of both model parameters and latent representations between the user and the server. However, their focus is traffic prediction using time-series user data, employing an encoder-decoder architecture in their user-side model to handle sequence data, rendering their model and algorithm incompatible with our setting. Nonetheless, we adapt their conceptual framework by allowing users to upload both the model parameter and the latent vector, with the server subsequently averaging the received local models and broadcasting the updated global model to all users. It's important to note that their approach simultaneously uploads the local model and the latent representation, significantly increasing the risk of privacy leakage. Prior research \cite{geiping2020inverting} has shown that adversaries could infer sensitive information from model parameters. 

\subsection{Experiment Results}

We first compare our method with the baseline. In Fig.~\ref{fig:compare_baseline_gcn}, we illustrate the training loss and testing accuracy of the split GCN model on the Cora, Citeseer, and Pubmed datasets. Note additional experiment results on Chameleon, Squirrel, and Wiki-CS datasets can be found in Fig.~\ref{fig:compare_gcn_csw} in Appendix~\ref{sec:extra_plot}. From the figure, we can observe that nFedGNN significantly outperforms the CNFGNN on all datasets. To show the effectiveness of regularization loss, we search the best value of $\lambda$ within the range $\{0.1, 1, 10, 100, 300 \}$ for Cora, Citeseer, and Pubmed, and the range $\{1, 10, 100, 500\}$ for Chameleon, Squirrel, and Wiki-CS. We select the best $\lambda$ for each dataset, and the optimal hyperparameter can be found in each image. In the figure, the performance improvement of nFedGNN with the regularization can be clearly observed. The best accuracy reaches $71.9\%$ for the GCN model on the Cora dataset. Here, we also plot the testing accuracy of the GCN model on these datasets in centralized training~\cite{kipf2016semi} as references. Although there is still a gap between nFedGNN and the centralized GNN model in testing accuracy, it is significantly reduced compared with the baseline and the no regularization counterpart. Note our goal is not to achieve the state-of-the-art but to demonstrate the effectiveness of the proposed approach. It is worth noting that CNFGNN needs to transmit the extra local model parameters between the users and server at every FL round, necessitating more communication resources. In Appendix~\ref{sec:extra_plot}, we demonstrate the comparison results for the GAT in Fig.~\ref{fig:compare_baseline_gat},~\ref{fig:compare_gat_cs}, similar trends and conclusions can be observed in these figures. In summary, nFedGNN has better performance, clearly proving the advantages of nFedGNN compared with baseline.

Subsequently, we investigate the influence of $\lambda$ on the model convergence. As shown in Figure~\ref{fig:lambda_gcn}, we display the testing accuracy and training loss under different magnitudes of $\lambda$. A general trend emerging from these figures is the testing accuracy increases when $\lambda$ is getting higher. However, when $\lambda$ exceeds some threshold value, the testing accuracy starts to decrease. For the training loss, a higher $\lambda$, the value is larger. Another observation is the convergence speed of testing accuracy is slower with higher $\lambda$, e.g., it takes more rounds for the testing accuracy to achieve the plateau. We conclude that with a higher value of $\lambda$, the training task is more ``difficult'', and training requires more time to converge. By properly determining the value of $\lambda$, the learned models' performance can be improved. In Appendix~\ref{sec:extra_plot}, we also illustrate the impact of $\lambda$ on model convergence for the GAT in Fig.~\ref{fig:lambda_gat}. Similar observations can be found in these figures. Note that due to the stochastic nature of the learning process, a few curves may not match the above observations. However, the general trend is clear, and the conclusions are reliable. 

\subsection{Communication Cost Comparison}
\begin{table}[t]
  \caption{Communication cost of GCN and GAT on different datasets (in MB).}
  \label{compare-table}
  \centering
    \begin{tabular}{|c|c|c|c|c|}
    \midrule
    \textbf{Method}  & \multicolumn{2}{c|}{\textbf{nFedGNN} } & \multicolumn{2}{c|}{\textbf{CNFGNN}} \\
    \midrule
               & GCN  & GAT            & GCN  & GAT  \\
    \midrule
    Cora       & 34.06   & 264.45              & $4.74 \times 10^4$  & $3.79 \times 10^5$ \\
    Citeseer   & 40.61   & 324.90              & $1.50 \times 10^5$  & $1.20 \times 10^6$ \\
    Pubmed     & 240.69  & $1.93 \times 10^3$  & $1.21 \times 10^5$  & $9.64 \times 10^5$  \\
    Chameleon  & 27.80   & 222.36              & $6.47 \times 10^4$  & $5.17 \times 10^5$ \\
    Squirrel   & 63.49   & 507.91              & $1.33 \times 10^5$  & $1.06 \times 10^6$ \\
    Wiki-CS    & 142.83  & $1.14 \times 10^3$  & $4.31 \times 10^4$  & $3.44 \times 10^5$ \\
    \bottomrule
\end{tabular}
\end{table} 

We count the total communication cost (in MB) of both nFedGNN and CNFGNN for the GCN model and the GAT models in Table \ref{compare-table}. This includes the cumulative communication cost spanning 200 training rounds for all users within the graph, with the data transmitted in float32 format. For nFedGNN, we record the size of the latent feature vector. For CNFGNN, we record the size of the latent feature vector and the number of user-side model parameters. In all cases, nFedGNN consumes less communication resources than CNFGNN. Moreover, the total transmitted data size is very small, verifying the conclusion that communication cost is not the bottleneck for our method.

\section{Conclusion}\label{sec:con}
We introduce nFedGNN, a novel federated learning algorithm for GNNs where each user only has access to the data of a single node. Our method provides an opportunity to utilize the distributed graph data without compromising user privacy. We compare nFedGNN with nFedGNN over multiple datasets and demonstrate the advantage of our method. Despite our extensive efforts, we recognize the limitations in the scope of our study. For example, we have not explored the training on larger datasets and inductive frameworks like GraphSage~\cite{hamilton2017inductive}. In the future, we will extend nFedGNN to adopt partial user participation and to provide rigorous privacy protection (e.g., differential privacy). 



\printbibliography

\newpage
\appendix

\noindent\rule{\textwidth}{1pt}
\begin{center}
\vspace{7pt}
{\Large  Appendix}
\end{center}
\noindent\rule{\textwidth}{1pt}

\section{Related Work}\label{sec:related}

\textbf{Federated Learning.} FL has drawn great attention recently due to its benefits in privacy and communication efficiency \cite{thapa2020splitfed}. The goal of FL is to allow multiple users to collaboratively train a model under the orchestration of a central server without sharing their raw data \cite{kairouz2019advances}. In conventional FL algorithms such as FedAvg \cite{mcmahan2017communication}, model training takes place on the user side, and only the model or model update is transferred between the user and a server. An aggregator on the server subsequently aggregates the models from the users for the next training round. In FL, the user's raw data is never shared. 

One branch of FL is vertical federated learning (VFL) \cite{wucoupled, wu2020privacy}. In conventional FL \cite{mcmahan2017communication}, the user has different data samples, while the samples from different users share the same feature space. In VFL, the user owns a common sample space, but disparate feature spaces. The first few layers of the model are partitioned width-wise to accommodate different feature spaces. Each user owns a portion of the whole model corresponding to their feature spaces. The user who possesses the labels holds the remaining layers of the model and coordinates the training process.

Another relevant notion of FL is split learning \cite{gupta2018distributed,vepakomma2018split}. Specifically, the model is partitioned layer-wise between the users and the server. Only the intermediate results, e.g., the activation maps and related gradients, are exchanged between the user and server during training. As opposed to FL, split learning requires users to communicate with the server every iteration, incurring heavy communication overhead and high training latency. 

\textbf{GCN in Distributed Setting.} There has been an increasing research interest in training GNN in a distributed setting. Most of the work focuses on either graph-level FL, where small graphs are distributed among multiple parties \cite{he2021fedgraphnn,xie2021federated}, or subgraph-level FL, where each party holds a sub-graph of the whole large graph \cite{he2021fedgraphnn,zhang2021subgraph,baek2022personalized,wang2020graphfl,scardapane2020distributed}.

Specifically, Baek et al.\cite{baek2022personalized} propose a personalized sub-graph FL algorithm, which allows each user to train a personalized model by selectively sharing knowledge across users. Zhang et al. \cite{zhang2021subgraph} trains a missing neighbor generator to handle the missing edges of the subgraphs across users. Wang et al. \cite{wang2020graphfl} employs the model-agnostic meta-learning (MAML) approach to tackle the Non-IID distributed data for the graph-based semi-supervised node classification task. Hu et al. \cite{hu2022fedgcn} suggested an online adjustable attention mechanism to aggregate the local models. These prior works mainly focus on the setting that each user holds a graph/sub-graph, where a local GNN model could be independently trained and serves as a base model for FedAvg.

In this paper, we consider a more challenging case where each user represents only one node in the graph and does not have access to other's data. It is, in fact, a special but the most challenging case of subgraph-level FL and could not be addressed by any of the prior work. The closest work to ours is \cite{meng2021cross}. They consider the spatiotemporal dynamic modeling tasks and propose an approach that explicitly encodes the underlying graph structure using GNNs. Their approach is specifically crafted to fit the problem and may be unsuitable for general GNN tasks. 

\section{General Model Splitting Strategy}\label{app:mps}

Generally, GNN can be described within the Message Passing Neural Networks (MPNN) framework \cite{gilmer2017neural}. As shown in (\ref{equ:messpass}), the hidden state of each node $\bm{h}^k_{i}$ relies on the message from their neighboring nodes. Thus, user $i$ requires $\bm{h}^k_{\mathcal{N}_i}$ from its neighbor users to compute $\bm{h}^k_{i}$. This leads to excessive privacy leakage. Therefore, layer-wise splitting for GNN is infeasible under the node-level FL setting. Alternatively, we seek to split the first layer of the GNN model. However, the conventional GNN layer follows the message-passing-then-encoding protocol. Directly splitting a GNN layer results in the message aggregation function $\textbf{Agg}_0(\cdot)$ located on the user side. To deal with this problem, we reformulated the GNN layer as an encoding-then-message-passing process:
\begin{equation}\label{equ:messpass_new}
    \begin{split}
        \bar{\bm{h}}^k_{i} & :=  \textbf{Update}_k\left(\bm{h}^{k}_{i}\right),\\
        \bm{h}^{k+1}_{i} & := \textbf{Agg}_k\left(\{\bar{\bm{h}}^k_{j}, \forall j \in \mathcal{N}_i\}\right),
    \end{split}
\end{equation}
Note (\ref{equ:messpass_new}) usually results in the same model as message-passing-then-encoding. Then we can split the first GNN layer into two parts: the update function $\textbf{Update}_0 (\cdot)$ on the user side and the message aggregation function $\textbf{Agg}_0(\cdot)$ on the server side. Taking the split GCN model~\ref{equ:spit-gcn} as an example, we can find the first layer's hidden state update function $\textbf{Update}_k (\cdot)$ is a linear projection with parameter $W^{(0)}$, and the message aggregation function $\textbf{Agg}_k(\cdot)$ is the matrix multiplication with normalized adjacency matrix $\tilde{A}$. Switching the order of the $\textbf{Update}_k (\cdot)$ and $\textbf{Agg}_k(\cdot)$ implies the different matrix multiplication orders. The associativity property of matrix multiplication immediately proves the same GCN model after reformulation.

\section{Dataset statistics}\label{sec:data_stat}
Cora, Citeseer, and Pubmed serve as citation graphs in which nodes symbolize scientific publications and edges denote citation relationships. Wiki-CS is a dataset grounded in Wikipedia, where nodes represent articles related to Computer Science, and edges signify hyperlinks between them. Similarly, Squirrel and Chameleon are networks comprised of page-to-page connections where nodes are articles from English Wikipedia, and edges depict mutual links between these articles. A summary detailing the number of nodes, edges, and labels for these datasets is provided in Table~\ref{data-table}.
\begin{table}
  \caption{Dataset statistics}
  \label{data-table}
  \centering
  \begin{tabular}{|c|c|c|c|c|c|}
    \midrule
    Dataset    & Nodes    & Edges    & Classes    & Features      & Label Rate     \\
    \midrule
    Cora      & 2,708    & 10,556   & 7          & 1,433          & 5.2 $\%$        \\
    Citeseer  & 3,327    & 9,228    & 6          & 3,703          & 3.6 $\%$         \\
    Pubmed    & 19,717   & 88,651   & 3          & 500            & 0.3 $\%$       \\
    Wiki-CS   & 11,701   & 431,726  & 10         & 300            & 5 $\%$       \\
    Chameleon   & 2277   & 36101  & 5         & 300            & 50.0 $\%$       \\
    Squirrel   & 5201   & 217073  & 5         & 2089            & 50.0 $\%$       \\
    \bottomrule
  \end{tabular}
\end{table} 

\section{GAT Splitting Strategy}\label{sec:gat_split}
A GAT layer comprises a learnable weight matrix $W\in \mathbb{R}^{F^{\prime}\times F}$ and a shared attention mechanism $a:\mathbb{R}^{F^{\prime}} \times \mathbb{R}^{F^{\prime}} \rightarrow \mathbb{R}$, where $F$ and $F^{\prime}$ are the number of input and output of the GAT layer. Note that $a$ is used to compute the \emph{attention coefficient} $e_{ij}$, which further depends on the feature vectors of nodes $i$ and $j$. Therefore, the user-side model contains the trainable weight matrix $W$, while the shared attention mechanism $a$ should be located on the server side. At each training round, the user-side model first transforms the local feature vector $x_i$ into $x_i^\prime = W_i x_i$, which is uploaded to the server later. The server uses $x_i^\prime$ to compute the attention coefficients $e_{ij}=a(x_i^\prime, x_j^\prime)$ and update the model. Moreover, the GAT model usually employs the multi-head attention mechanism to stabilize the training process. To achieve this, we let the user have multiple weight matrices. Each weight matrix transforms the input feature independently: $x_{i,l}^\prime = W_i^{l} x_{i}^\prime$, where $W_i^l$ is the weight of $l$-th attention mechanism on $i$-th user. Subsequently, the user uploads a set of latent features $\{x_{i,l}' \}_{l=1}^{L}$ to the server, where $L$ is the number of attention mechanisms. Finally, the output of the first GAT layer is
\begin{equation}
x_{i}^{''} = \concat_{l=1}^L \sigma\left( \sum_{j\in\mathcal{N}_i} \alpha_{ij}^l x_{i,l}' \right),
\end{equation}
where $\sigma$ is the activation function and $\alpha_{ij}^l=\text{softmax}(e_{ij}^{l})$ is the normalized attention coefficients computed by the $l$-th attention mechanism. 

The second GAT layer is the same as that in the GAT model for the task on the Cora dataset in \cite{velivckovic2017graph}. For the multi-head GAT model, the regularization term of nFedGNN has two variations. The first approach treats every element in $\{x_{i,l}' \}_{l=1}^{L}$ equally and computes the regularization for each $x_{i,l}'$ independently. Alternatively, we first compute the average latent representation $\bar{x}_{i}'=\frac{1}{L}\sum_{l=1}^{L} x_{i,l}'$, and then obtain the regularization term based on $\bar{x}_{i}'$. In this work, we adopt the first scheme. The performance of the second scheme and the comparison between the two variations are beyond the scope of this work, and we left it for future exploration.

\section{Extra Experiment Result}\label{sec:extra_plot}
\begin{figure}[t]
\centering
    \subfloat[]{{\includegraphics[width=0.33\textwidth]{ {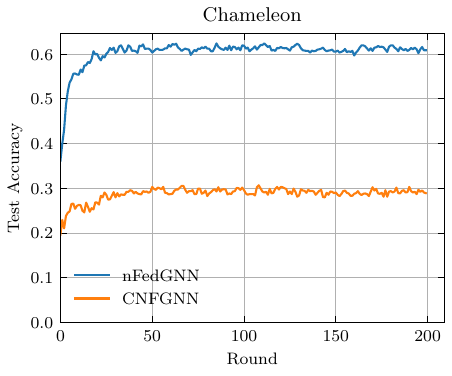}} }}
    \subfloat[]{{\includegraphics[width=0.33\textwidth]{ {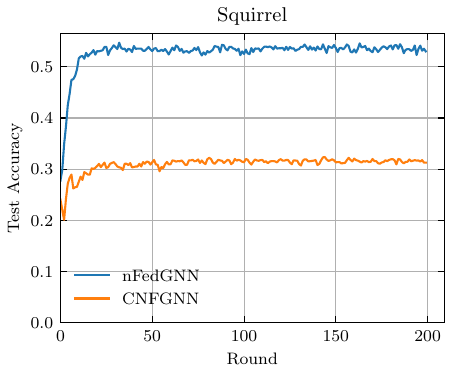}} }}
    \subfloat[]{{\includegraphics[width=0.33\textwidth]{ {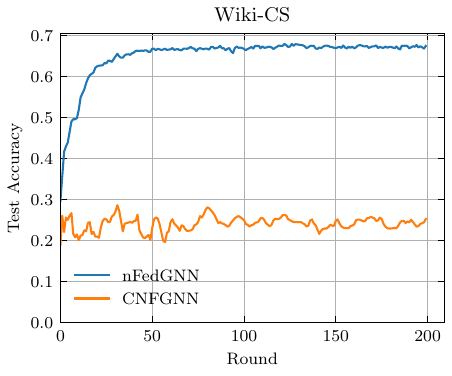}} }} \\
    \subfloat[]{{\includegraphics[width=0.33\textwidth]{ {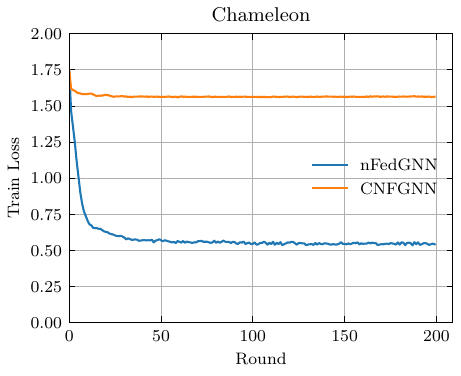}} }}
    \subfloat[]{{\includegraphics[width=0.33\textwidth]{ {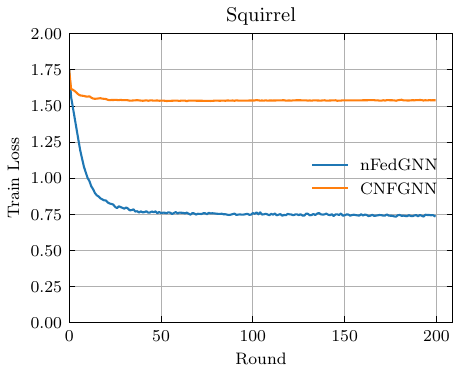}} }}
    \subfloat[]{{\includegraphics[width=0.33\textwidth]{ {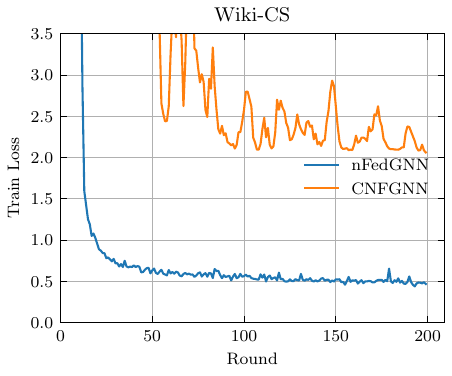}} }}
    \caption{nFedGNN vs CNFGNN for GCN on the Chameleon, Squirrel, and Wiki-CS datasets.}%
    \label{fig:compare_gcn_csw}%
\end{figure}

\begin{figure}[t]
\centering
    \subfloat[]{{\includegraphics[width=0.33\textwidth]{ {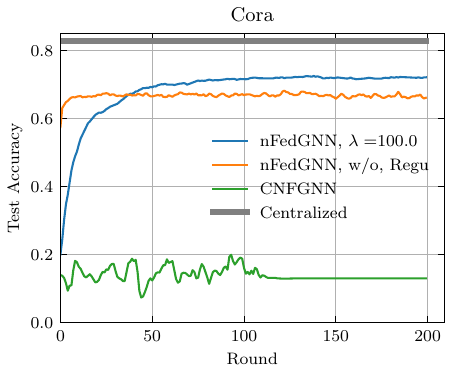}} }}
    \subfloat[]{{\includegraphics[width=0.33\textwidth]{ {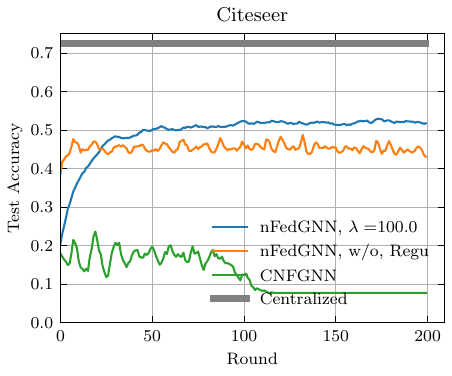}} }}
    \subfloat[]{{\includegraphics[width=0.33\textwidth]{ {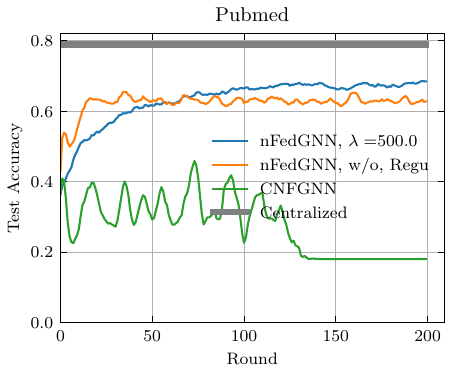}} }} \\
    \subfloat[]{{\includegraphics[width=0.33\textwidth]{ {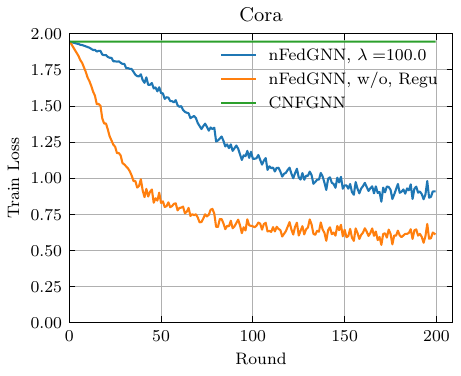}} }}
    \subfloat[]{{\includegraphics[width=0.33\textwidth]{ {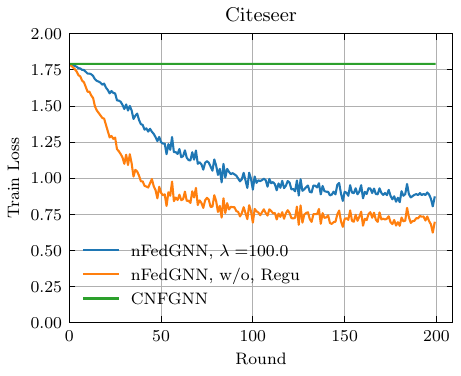}} }}
    \subfloat[]{{\includegraphics[width=0.33\textwidth]{ {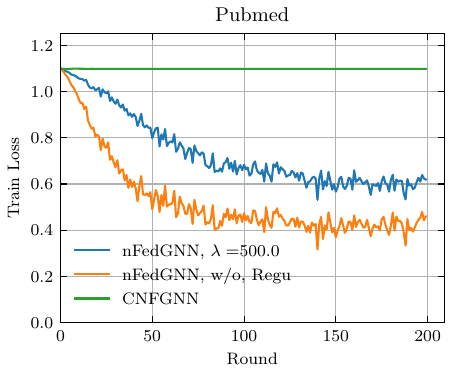}} }}
    \caption{nFedGNN vs CNFGNN for GAT on the Cora, Citeseer, and Pubmed datasets.}%
    \label{fig:compare_baseline_gat}%
\end{figure}

\begin{figure}[t]
\centering
    \subfloat[]{{\includegraphics[width=0.33\textwidth]{ {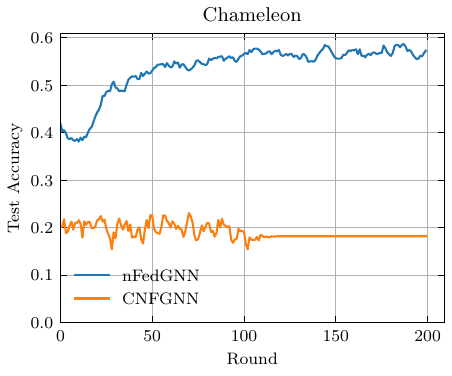}} }}
    \subfloat[]{{\includegraphics[width=0.33\textwidth]{ {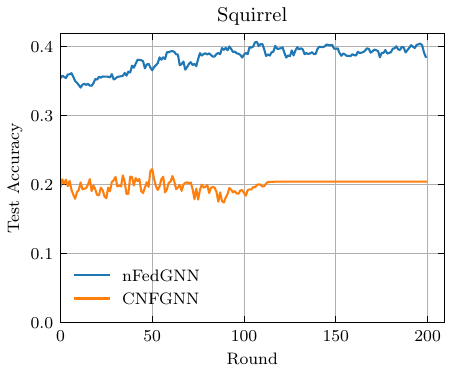}} }}\\
    \subfloat[]{{\includegraphics[width=0.33\textwidth]{ {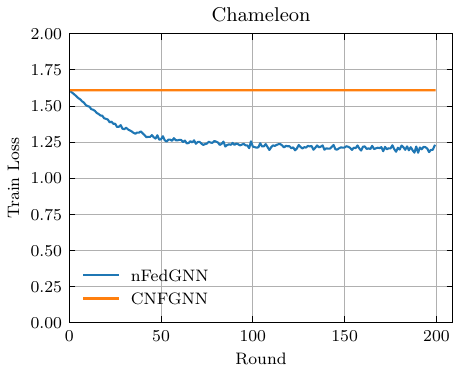}} }}
    \subfloat[]{{\includegraphics[width=0.33\textwidth]{ {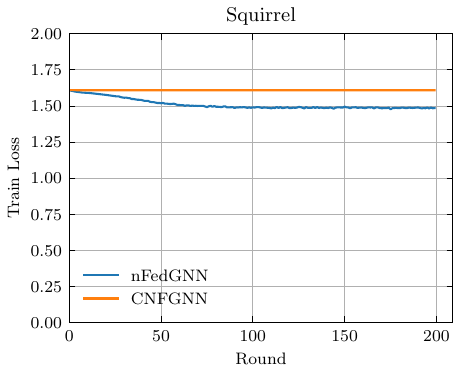}} }}
    \caption{nFedGNN vs CNFGNN for GAT on the  Chameleon and Squirrel datasets.}%
    \label{fig:compare_gat_cs}%
\end{figure}

\begin{figure}[t]
\centering
    \subfloat[]{{\includegraphics[width=0.33\textwidth]{ {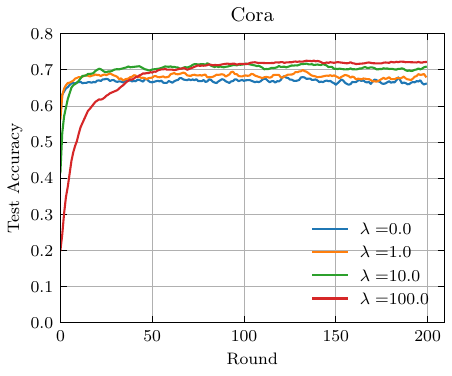}} }}
    \subfloat[]{{\includegraphics[width=0.33\textwidth]{ {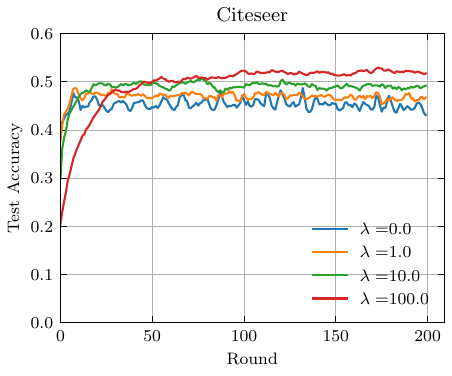}} }}
    \subfloat[]{{\includegraphics[width=0.33\textwidth]{ {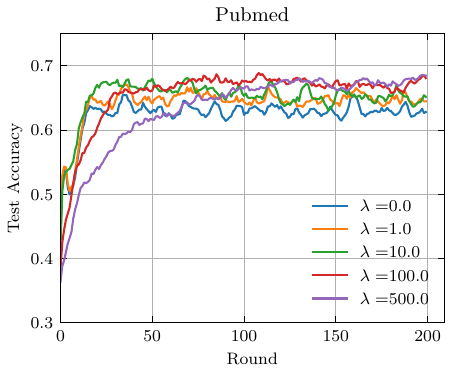}} }} \\
    \subfloat[]{{\includegraphics[width=0.33\textwidth]{ {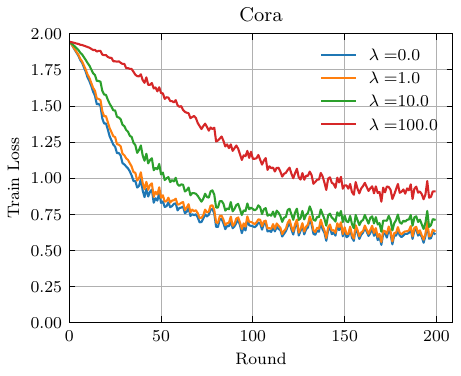}} }}
    \subfloat[]{{\includegraphics[width=0.33\textwidth]{ {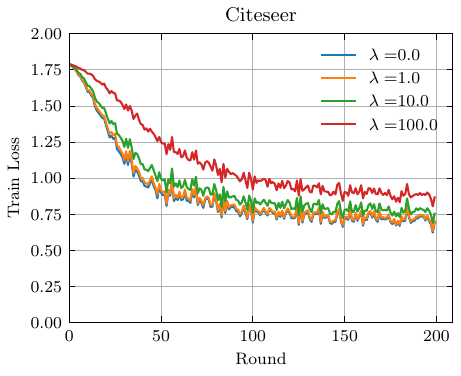}} }}
    \subfloat[]{{\includegraphics[width=0.33\textwidth]{ {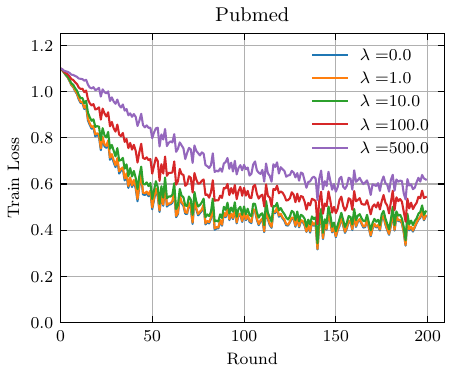}} }}
    \caption{Impact of $\lambda$ for GAT on the Cora, Citeseer, and Pubmed datasets.}%
    \label{fig:lambda_gat}%
\end{figure}

\end{document}